\pdfoutput=1

\documentclass[11pt]{article}
\usepackage[]{naacl2021}

\usepackage{times}
\usepackage{latexsym}

\usepackage{siunitx}
\usepackage{multirow}
\usepackage{microtype}
\usepackage{graphicx}
\usepackage{subfigure}
\usepackage{booktabs} 
\usepackage{hyperref}
\usepackage{amsmath}

\usepackage{amsbsy}
\usepackage{bm}
\usepackage{float}
\usepackage{lipsum}
\usepackage{hhline}
\usepackage{microtype}
\usepackage[T2A,T1]{fontenc}

\usepackage[utf8]{inputenc}
\usepackage[russian,english]{babel}

\title{Neural Machine Translation without Embeddings}

\author{Uri Shaham$^\diamondsuit$ \quad Omer Levy$^\diamondsuit$$^\spadesuit$\\
\\ $^\diamondsuit$The Blavatnik School of Computer Science, Tel Aviv University
\\ $^\spadesuit$Facebook AI Research
}

\date{}

\begin{document}
\maketitle

\begin{abstract}
Many NLP models operate over sequences of subword tokens produced by hand-crafted tokenization rules and heuristic subword induction algorithms.
A simple universal alternative is to represent every computerized text as a sequence of bytes via UTF-8, obviating the need for an embedding layer since there are fewer token types (256) than dimensions.
Surprisingly, replacing the ubiquitous embedding layer with one-hot representations of each byte does not hurt performance; experiments on byte-to-byte machine translation from English to 10 different languages show a consistent improvement in BLEU, rivaling character-level and even standard subword-level models.
A deeper investigation reveals that the combination of embeddingless models with decoder-input dropout amounts to token dropout, which benefits byte-to-byte models in particular.\footnote{Our code is publicly available at: \url{https://github.com/UriSha/EmbeddinglessNMT}}
\end{abstract}
\section{Introduction}

Neural NLP models often operate on the subword level, which requires language-specific tokenizers \cite{koehn-etal-2007-moses,adler-elhadad-2006-unsupervised} and subword induction algorithms, such as BPE \cite{Sennrich_2016,kudo-2018-subword}.
Instead, working at the byte level by representing each character as a variable number of Unicode (UTF-8) bytes, does not require any form of preprocessing, allowing the model to read and predict every computerized text using a single vocabulary of 256 types.
While previous work found that byte-level models tend to underperform models based on subword tokens \cite{wang2019neural},
byte-based models exhibit an interesting property: their vocabulary is smaller than the number of latent dimensions ($256 < d$).
In this work, we demonstrate that this property allows us to \textit{remove} the input and output embedding layers from byte-to-byte translation models, and in doing so, \textit{improve} the models' performance consistently.

We replace the dense trainable embedding matrix with a fixed one-hot encoding of the vocabulary as the first and last layers of a standard transformer model.
Machine translation experiments on 10 language pairs show that byte-to-byte models \textit{without} an embedding layer achieve higher BLEU scores than byte-based models with parameterized embeddings (+0.5 on average), thus closing the performance gap with subword and character models. 
We observe this result consistently throughout a wide variety of target languages and writing systems.

The fact that removing parameters improves performance is counter-intuitive, especially given recent trends in machine learning that advocate for increasingly larger networks.
We further investigate why embeddingless models yield better results and find implicit token dropout (commonly referred to as ``word dropout'') as the main source of that boost.
While prior work shows that randomly masking tokens from the decoder input can improve the performance of language generation models \cite{bowman-etal-2016-generating}, we find that this effect is amplified when operating at the byte level.
Overall, our results suggest that, even without additional parameters, byte-based models can compete and potentially outperform subword models, but that they may require alternative optimization techniques to achieve that goal.

\begin{figure*}[ht]
\small
\centering
\begin{tabular}{lc@{}|@{}c@{}|@{}c@{}|@{}c@{}|@{}c@{}|@{}c@{}|@{}c@{}|@{}c@{}|@{}c@{}|@{}c@{}|@{}c@{}|@{}c@{}|@{}c@{}|@{}c@{}|@{}c@{}|@{}c@{}|@{}c@{}|@{}c@{}|@{}c@{}|@{}c@{}|@{}c@{}|@{}c}
\toprule
\textbf{Original Text}  & \multicolumn{22}{c}{\begin{otherlanguage*}{russian}Будь здоров.\end{otherlanguage*}}\\
\textbf{Subwords (BPE)}  & \multicolumn{4}{@{}c@{}|}{\begin{otherlanguage*}{russian}Бу\end{otherlanguage*}@} & \multicolumn{5}{@{}c@{}|}{\begin{otherlanguage*}{russian}дь\end{otherlanguage*}} & \multicolumn{6}{@{}c@{}|}{\begin{otherlanguage*}{russian}здо\end{otherlanguage*}@} & \multicolumn{6}{@{}c@{}|}{\begin{otherlanguage*}{russian}ров\end{otherlanguage*}} & .\\
\textbf{Characters} & \multicolumn{2}{@{}c@{}|}{\begin{otherlanguage*}{russian}Б\end{otherlanguage*}} & \multicolumn{2}{@{}c@{}|}{\begin{otherlanguage*}{russian}у\end{otherlanguage*}} & \multicolumn{2}{@{}c@{}|}{\begin{otherlanguage*}{russian}д\end{otherlanguage*}} & \multicolumn{2}{@{}c@{}|}{\begin{otherlanguage*}{russian}ь\end{otherlanguage*}} &   & \multicolumn{2}{@{}c@{}|}{\begin{otherlanguage*}{russian}з\end{otherlanguage*}} & \multicolumn{2}{@{}c@{}|}{\begin{otherlanguage*}{russian}д\end{otherlanguage*}} & \multicolumn{2}{@{}c@{}|}{\begin{otherlanguage*}{russian}о\end{otherlanguage*}} & \multicolumn{2}{@{}c@{}|}{\begin{otherlanguage*}{russian}р\end{otherlanguage*}} & \multicolumn{2}{@{}c@{}|}{\begin{otherlanguage*}{russian}о\end{otherlanguage*}} & \multicolumn{2}{@{}c@{}|}{\begin{otherlanguage*}{russian}в\end{otherlanguage*}} & .  \\
\textbf{Bytes (UTF-8)} & \ D0 \  & \ 91 \ & \ D1 \ & \ 83 \ & \ D0 \ & \ B4 \ & \ D1 \ & \ 8C \ & \ 20 \ & \ D0 \ & \ B7 \ & \ D0 \ & \ B4 \ & \ D0 \ & \ BE \ & \ D1 \ & \ 80 \ & \ D0 \ & \ BE \ & \ D0 \  & \  B2 \  &\  2E  \\
\bottomrule
\end{tabular}
\caption{Subword (BPE), character, and byte tokens of the string ``\begin{otherlanguage*}{russian}Будь здоров\end{otherlanguage*}.''
UTF-8 uses two bytes to represent each character in the Cyrillic script, making the byte sequence longer than the number of characters.}
\label{fig:tokenization}
\end{figure*}

\section{Byte Tokenization}
\label{sec:tokenization}

Modern software typically represents text using Unicode strings (UTF-8), which allows one to encode virtually any writing system using a variable number of bytes per token; English characters are typically represented by a single byte, with other writing systems taking two (e.g. Arabic), three (e.g. Chinese), or four (e.g. emojis) bytes per character.
By treating each byte as a separate token, we can encode any natural language text using a single universal vocabulary of only 256 token types.
Moreover, byte tokenization obviates the need for any heuristic preprocessing, such as splitting spaces, punctuation, and contractions.
Figure~\ref{fig:tokenization} illustrates subword, character, and byte tokenization.

\section{Embeddingless Model}
\label{sec:model}

Our model is based on the original transformer encoder-decoder \cite{vaswani2017attention} with one main difference: we eliminate the input and output token embedding layers.
These layers typically use a common parameter matrix $E \in R^{|V| \times d}$ that contains a $d$-dimensional embedding vector for each source and target vocabulary item in $V$.\footnote{One could argue that the first layer of each transformer stack (the key, query, and value matrices) qualify as some form of multi-head multi-purpose embedding layer, where each token type is effectively represented by $3h$ different vectors ($h$ being the number of attention heads) in the encoder and $3h$ additional vectors in the decoder. This is very different from the standard notion of embeddings, where each token type has a universal representation that can be shared across the encoder input, decoder input, and decoder output.}

Instead, we use a fixed one-hot representation of our byte vocabulary.
For instance, the character ``R'' could be represented as a vector with 1 at dimension 82 and 0 elsewhere. 
Since it is standard practice to use representations of more than 256 dimensions, every possible byte can be represented by such one-hot vectors.
To predict the next token for a decoder input of $n$ tokens, we take the output of the last transformer decoder layer, $Y \in R^{n \times d}$, and apply a softmax across each vector's dimensions. Formal expressions of the input and output of our model are detailed in Figure \ref{fig:embeddingless}.

Omitting the embedding layer reduces the number of parameters by a factor of $O(|V| \cdot d)$.\footnote{For subword tokenization, this accounts for a significant portion of the parameter budget, but for byte-based models the added parameter cost is negligible.} 
We do add a total of 3 parameters to scale the encoder and decoder's (one-hot) inputs and the decoder's output (before the softmax). 
We initialize all three with $\sqrt{d}$, akin to the constant scaling factor typically applied to the input embedding layer in transformers.
Despite the reduction in model size, memory consumption increases when working on longer sequences, since the space complexity of transformers is $O(n^2 + n \cdot d)$.
In our case, $d$ (512) is typically larger than $n$ (see Table \ref{tab:data}), entailing an increase in memory consumption that is roughly linear in the sequence length $n$, and a similar decrease in processing speed when compared to character and subword models.

In addition to replacing the embedding layers, we also remove the dropout layers on the encoder input and decoder output, since zeroing out entries of one-hot vectors is equivalent to randomly masking out input tokens or deleting significant parts of the model's predicted distribution.
The dropout on the decoder input (prefix of the target fed with teacher forcing) remains intact at this point and is applied throughout our main experiments. 
Further analysis shows that decoder input dropout is in fact a significant source of performance gains, which we further investigate in Section~\ref{sec:analysis}.

\begin{figure}[t]
\small
\centering
\begin{tabular}{ccc}
\toprule
      & \textbf{Original} & \textbf{Embeddingless} \\
\midrule
\textbf{Input}  & $X E + P_n$ & $X + P_n$ \\
\textbf{Output} & $\text{softmax}_{|V|} \left( Y E^\top \right)$ & $\text{softmax}_d \left( Y \right)$ \\
\bottomrule
\end{tabular}
\caption{The main differences between the original encoder-decoder model and the new embeddingless model. $X \in R^{n \times |V|}$ is the one-hot representation of $n$ input tokens (bytes); $P_n$ are the positional embeddings up to length $n$.} 
\label{fig:embeddingless}
\end{figure}
\section{Experiments}
We train byte-tokenized embeddingless models for machine translation and compare them to standard byte, character, and subword-based models on a diverse set of languages.
We adopt a standard experimental setup that was designed and tuned for the subword baseline and limits our hyperparameter tuning to dropout probabilities.

\paragraph{Datasets}
We use the IWSLT\footnote{All languages used the IWSLT2014 data except for Vietnamese (IWSLT2015) and Japanese (IWSLT2017).} datasets of English TED talks translated into other languages \cite{cettolo2014report}, selecting 10 additional languages with varying characteristics\footnote{While in this work we prioritized language and writing system diversity, there is room to test embedingless models on larger datasets in future work.} (see Table~\ref{tab:data}).
For each one, we train translation models from English to the target language (the original direction of translation ), and also in the opposite direction for completeness.
We clean the training data for every language pair by first removing sentences longer than 800 bytes, and then the sentences with the largest byte-length ratio between source and target such that we remove a total of 5\% of the training examples.




\begin{table}[t]
\small
\centering
\begin{tabular}{@{}llcccc@{}}
\toprule
\multirow{2}{*}{\textbf{Language}} & \multirow{2}{*}{\textbf{ID}} &  \multirow{2}{*}{\textbf{\#Sentences}} & \multicolumn{3}{@{}c}{\textbf{Average length}} \\
& &  & \textbf{BPE} & \textbf{Char} & \textbf{Byte} \\
\midrule
Chinese & zh & 166k & 20.9  &  32.4  &  90.1 \\
Spanish & es & 167k & 25.4   &  100.2  &  100.2 \\
Arabic & ar & 166k & 24.4 &   79.3 &    142.2 \\
Russian & ru & 164k & 26.3  &  93.9  &  169.7 \\
German & de & 159k &  26.6  &  106.5 &  107.9 \\
Japanese & ja & 215k & 20.9  &   42.4  &  115.3 \\
Turkish & tr & 143k & 24.1 &   93.6  &  102.0 \\
Vietnamese & vi & 124k & 26.9  &  99.8  &  132.5 \\
Farsi & fa & 100k & 27.4  &  93.1  &   165.9 \\
Hebrew & he & 171k & 23.0 &   72.8  &  129.2 \\
English & en & - & 25.6  &  97.0  &  97.1  \\
\bottomrule
\end{tabular}
\caption{Languages from the IWSLT dataset, along with the number of sentence pairs in the training set and the average sequence length per tokenization method.}
\label{tab:data}
\end{table}

\paragraph{Baselines}
In addition to the byte-based embeddingless transformer, we train standard transformer encoder-decoder models as baselines, each one using a different tokenization scheme: subword, character, and byte. For subword tokenization, we apply the Moses tokenizer \cite{koehn-etal-2007-moses} followed by BPE \cite{Sennrich_2016}. Both character and byte tokenizations apply no additional preprocessing at all and include whitespaces as valid tokens.

\paragraph{Hyperparameters}
The code for our model and baselines is based on Fairseq \cite{ott-etal-2019-fairseq} implementation of the transformer encoder-decoder model. 
During preprocessing we use 10,000 merging steps when building the BPE vocabulary for every language pair.
The vocabularies and embeddings are always shared among source and target languages.
In every transformer we use 6 encoder and decoder layers, 4 attention heads, a hidden dimension of 512, and a feed-forward dimension of 1024. 
We optimize with Adam \cite{kingma2014adam}, using the inverse square root learning rate scheduler with 4000 warmup steps and a peak learning rate of \num{5e-4}, label smoothing of 0.1, and weight decay of \num{1e-4}.
We train each model for 50k steps and average the top 5 checkpoints according to the validation loss.
We tune dropout (0.2 or 0.3) on the validation set.
We set the batch size according to a maximum of 64,000 bytes per batch, which controls for the number of batches per epoch across different tokenization methods.

\paragraph{Evaluation}
We evaluate our models using SacreBLEU, case-sensitive, with the 13a tokenizer for all languages except Chinese (ZH tokenizer) and Japanese (MeCab tokenizer).
We use the raw text as the reference for all of our experiments, instead of using the default tokenized-detokenized version, which normalizes the text and gives an artificial advantage to text processed with Moses.
\section{Results}
\label{sec:results}

Table \ref{tab:results} shows our experiments' results. Every row describes the test BLEU scores of our model and the three baselines trained on a different language pair. We discuss the implications of these below. 

\begin{table}[t]
\begin{center}
\begin{small}
\begin{tabular}{@{}cccccc@{}}
\toprule
\multicolumn{2}{@{}c}{\textbf{Benchmark}} & \multicolumn{3}{c}{\textbf{Embedding-based Models}} & \textbf{Embed-less} \\
\textbf{Src} & \textbf{Tgt} &  \textbf{Subword}  & \textbf{Char} &  \textbf{Byte}  & \textbf{Byte} \\
\midrule
en & zh  & 19.9& 20.8  & 20.2 & \textbf{21.0}\\
en & es  & \textbf{36.8}& 36.3  & 36.3 & \textbf{36.8}\\
en & ar  & 12.5& 12.5  & 12.3 & \textbf{12.9}\\
en & ru  & 18.1& 17.6  & 17.4 & \textbf{18.2}\\
en & de  & \textbf{29.4}& 28.6  & 28.7 & 29.1\\
en & ja  & 12.0 & 12.5  & 12.5 & \textbf{13.1}\\
en & tr  & 13.6& 13.7  & 13.8 & \textbf{14.1}\\
en & vi  & \textbf{29.7}& 28.2  & 28.0 & 28.7\\
en & fa  & 11.5& 11.7  & 12.0 & \textbf{12.1}\\
en & he  & 26.1& \textbf{26.9}  & 26.4 & 26.7\\
  \midrule
zh  & en &   \textbf{16.8} & 16.6  & 15.6 & 16.1\\
es  & en &   \textbf{39.6} & 38.5  & 38.4 & 38.8\\
ar  & en &   \textbf{31.5} & 30.2  & 30.3 & 30.8\\
ru  & en &   \textbf{22.7} & 21.9  & 22.0 & 22.0\\
de  & en &   \textbf{35.4} & 34.0  & 34.1 & 34.5\\
ja  & en &   \textbf{13.1} & 12.6  & 11.4 & 12.2 \\
tr  & en &   \textbf{23.3} & 22.5  & 22.3 & \textbf{23.3}\\
vi  & en &   \textbf{26.8} & 25.0  & 24.7 & 25.3\\
fa  & en &   \textbf{23.5} & 22.4  & 22.1 & 22.6\\
he  & en &   \textbf{37.8} & 36.9  & 37.0 & 37.4 \\
\bottomrule
\end{tabular}
\end{small}
\end{center}
\caption{Test BLEU scores of the baseline and embeddingless models on the IWSLT dataset. \label{tab:results}} 
\end{table}

\paragraph{Are embeddings essential?}
The results show that it is indeed possible to train embeddingless machine translation models that perform competitively.
The performance gaps between models with different tokenization schemes are relatively small.
Except for Vietnamese, the difference between the embeddingless model and the best embedding-based model is always under 1 BLEU.

In the most controlled setting, where we compare byte-based models with and without learnable embeddings, models \emph{without} embeddings consistently achieve higher BLEU scores in 19 of 20 cases (and an equal score for ru-en), with a boost of about 0.5 BLEU on average.
When compared to models based on character embeddings, the embeddingless byte-to-byte approach yields higher BLEU scores in 17 out of 20 cases, though the average difference is quite small in practice (0.3 BLEU).



\paragraph{Is subword tokenization superior to bytes or characters?}
Previous work in machine translation shows that subword models consistently outperform character or byte-based models \cite{gupta2019character,wang2019neural,gao-etal-2020-character}.
However, our results indicate that this is not necessarily the case.
When translating from English to a foreign language, the \textit{original} direction of the IWSLT dataset, embeddingless byte-to-byte models achieve performance that is equal or better than subword embedding models' in 8 out of 10 cases.
We observe a different trend when translating into English, where subword models surpass other models for every source language; the fact that Moses is a particularly good tokenizer for English -- and less so for other languages -- is perhaps related to this phenomenon.
Whereas prior work proposed closing the performance gap by adding layers to the basic architecture, under the assumption that character-based models lack capacity or expressiveness, our results show that actually \emph{removing} a component from the model can improve performance under certain conditions.
It is possible that character and byte-based transformer models encounter an optimization issue rather than one of capacity or expressivity.

\section{Analysis}
\label{sec:analysis}

Why does \textit{removing} the embedding matrix \textit{improve} the performance of byte-based models?
As mentioned in Section~\ref{sec:model}, the embeddingless models do not use dropout on the encoder input and decoder output, but do apply dropout on the decoder input while training.
Since the embeddingless decoder's inputs are fixed one-hot vectors, using dropout implicitly drops out complete tokens.
In prior work, token dropout (``word dropout'') has been shown to have a consistently positive effect \cite{bowman-etal-2016-generating}.
We, therefore, rerun our experiments while controlling for token dropout ($p=0.2$) to determine its effect on our results.

Table~\ref{tab:ablation} shows that decoder-side token dropout improves the performance of all models, with a larger impact on byte-based models and embeddingless models in particular.
This effect is largely consistent, with only 7 out of 160 cases in which token dropout decreased performance on the validation set.
We suspect that dropping out target tokens softens the effects of exposure bias by injecting noise into the ground-truth prefix.

Given the benefits of token dropout on the baseline models, we re-evaluate the results from Section~\ref{sec:results}, while allowing for token dropout as a potential hyperparameter.
Table~\ref{tab:results_mdi} shows that, when translating from the original English text to a foreign language, the different models perform roughly on par, with no single tokenization method dominating the others. 
Furthermore, byte-level models with and without embeddings achieve almost identical results.
In contrast, when translating in the opposite direction, subword models consistently outperform the other methods with an average gap of 0.76 BLEU from the next best model. Also, removing the embeddings from byte-based models decreases performance by an average of 0.45 BLEU when generating English.
This discrepancy might stem from artifacts of reverse translation, or perhaps from the English-centric nature of subword tokenization, which is based on Moses preprocessing and BPE. 
Overall, these results suggest that despite the greater number of parameters in subword models, character and byte models can perform competitively, but may require slightly different optimization techniques to do so.

\begin{table}[t]
\begin{center}
\begin{small}
\begin{tabular}{@{}lcccc@{}}
\toprule
 & \multicolumn{3}{c}{\textbf{Embedding-based Models}} & \textbf{Embed-less} \\
 &  \textbf{Subword}  & \textbf{Char} &  \textbf{Byte}  & \textbf{Byte} \\
\midrule
en$\rightarrow$xx  & +0.33 &  +0.53 & +0.42 & +0.62\\
xx$\rightarrow$en &  +0.69 & +0.67 & +0.92 & +0.83 \\
\bottomrule
\end{tabular}
\end{small}
\end{center}
\caption{The validation set performance gain of token dropout (0.2), averaged across languages and model dropout values.} 
\label{tab:ablation}
\end{table}

\begin{table}[t]
\begin{center}
\begin{small}
\begin{tabular}{@{}cccccc@{}}
\toprule
\multicolumn{2}{@{}c}{\textbf{Benchmark}} & \multicolumn{3}{c}{\textbf{Embedding-based Models}} & \textbf{Embed-less} \\
\textbf{Src} & \textbf{Tgt} &  \textbf{Subword}  & \textbf{Char} &  \textbf{Byte}  & \textbf{Byte} \\
\midrule
en & zh & 20.3 &  \textbf{21.2} & 20.8 & 21.0 \\
en & es & 36.7 &  \textbf{36.8} &  \textbf{36.8} &  \textbf{36.8} \\
en & ar & 12.7 &  \textbf{13.1} & 12.7 & 12.9 \\
en & ru &  \textbf{18.5} & 18.2 & 17.7 & 18.2 \\
en & de &  \textbf{29.8} & 29.3 & 29.2 & 29.1 \\
en & ja & 12.4 &  \textbf{13.1} & 12.5 &  \textbf{13.1} \\
en & tr & 13.9 & 14.3 &  \textbf{14.4} & 14.1 \\
en & vi &  \textbf{30.0} & 29.1 & 28.9 & 28.7 \\
en & fa & 11.5 &  \textbf{12.2} & 12.1 & 12.1 \\
en & he & 26.8 &  \textbf{27.1} &  \textbf{27.1} & 26.7 \\
\midrule
zh & en &  \textbf{17.3} & 17.2 & 16.3 & 16.1 \\
es & en &  \textbf{40.0} & 39.1 & 39.1 & 38.8 \\
ar & en &  \textbf{32.0} & 31.1 & 31.2 & 30.8 \\
ru & en &  \textbf{22.9} & 22.4 & 22.5 & 22.0 \\
de & en &  \textbf{35.6} & 34.9 & 35.0 & 34.5 \\
ja & en &  \textbf{13.5} & 12.8 & 12.3 & 11.2 \\
tr & en &  \textbf{24.3} & 23.3 & 23.7 & 23.3 \\
vi & en &  \textbf{27.4} & 25.9 & 25.9 & 25.3 \\
fa & en &  \textbf{24.5} & 23.2 & 23.3 & 22.6 \\
he & en &  \textbf{38.2} & 37.8 & 37.4 & 37.4 \\
\bottomrule
\end{tabular}
\end{small}
\end{center}
\caption{Test BLEU scores of the baseline and embeddingless models on the IWSLT dataset, when decoder-side token dropout is considered as a potential hyperparameter setting.} \label{tab:results_mdi}
\end{table}

\section{Related Work}

There is prior work on replacing language-specific tokenizers with more universal tokenization approaches. \citet{schutze-2017-nonsymbolic} shows how character n-gram embeddings can be effectively trained by segmenting text using a stochastic process. SentencePiece \cite{kudo-richardson-2018-sentencepiece} tokenizes raw Unicode strings into subwords using BPE \cite{Sennrich_2016} or unigram LM \cite{kudo-2018-subword}. Byte BPE \cite{wang2019neural} extends SentencePiece to operate at the byte level.
While this approach is indeed more language-agnostic than heuristic tokenizers, it does suffer from performance degradation when no pre-tokenization (e.g. splitting by whitespace) is applied.\footnote{\url{https://github.com/google/sentencepiece/blob/master/doc/experiments.md}}
Moreover, the assumption that subword units must be contiguous segments does not hold for languages with non-concatenative morphology such as Arabic and Hebrew.

Character and byte-based language models \cite{lee-etal-2017-fully, Al_Rfou_2019} treat the raw text as a sequence of tokens (characters or bytes) and do not require any form of preprocessing or word tokenization, and \citet{choe2019bridging} even demonstrate that byte-based language models can perform comparably to word-based language models on the billion-word benchmark \cite{chelba2013billion}.
Although earlier results on LSTM-based machine translation models show that character tokenization can outperform subword tokenization \cite{cherry-etal-2018-revisiting},
recent literature shows that the same does not hold for transformers \cite{gupta2019character,wang2019neural,gao-etal-2020-character}.
To narrow the gap, recent work suggests using deeper models \cite{gupta2019character} or specialized architectures \cite{gao-etal-2020-character}.
Our work deviates from this trend by \emph{removing} layers to improve the model.
This observation contests the leading hypothesis in existing literature -- that the performance gap results from reduced model capacity -- and suggests that the problem may be one of optimization.

\section{Conclusions}

This work challenges two key assumptions in neural machine translation models: the necessity of embedding layers, and the superiority of subword tokenization.
Experiments on 10 different languages show that, despite their ubiquitous usage, competitive models can be trained without any embeddings by treating text as a sequence of bytes. 
Our investigation suggests that different tokenization methods may require revisiting the standard optimization techniques used with transformers, which are primarily geared towards sequences of English subwords.



\section*{Acknowledgements}
This work was supported in part by Len Blavatnik and the Blavatnik Family foundation, the Alon Scholarship, and the Tel Aviv University Data Science Center.

\bibliography{anthology,references}
\bibliographystyle{acl_natbib}

\end{document}